\documentclass{article}

\usepackage{microtype}
\usepackage{graphicx}
\usepackage{subfigure}
\usepackage{booktabs} 
\usepackage{amsmath,amsfonts,amsthm}
\usepackage{mathrsfs}
\usepackage{cancel}

\usepackage{hyperref}

\usepackage[accepted]{icml2018}

\icmltitlerunning{}

\newcommand{\bz}{\mathbf{z}}
\newcommand{\br}{\mathbf{r}}
\newcommand{\bs}{\mathbf{s}}

\newcommand{\bphi}{\boldsymbol \phi}

\newcommand{\citeauthornum}[1]{\citeauthor{#1}~\citep{#1}}

\newcommand{\tar}{\pi}
\newcommand{\utar}{\pi^*}
\newcommand{\Real}{\mathbb{R}}

\newcommand{\cF}{\mathcal{F}}

\newcommand{\cQ}{\mathcal{Q}}
\newcommand{\dkl}{D_{\text{KL}}}
\newcommand{\dtv}{D_{\text{TV}}}
\newcommand{\Exp}{\mathbf{E}}

\DeclareMathOperator*{\argmin}{arg\,min}

\newtheorem{theorem}{Theorem}
\newtheorem{proposition}{Proposition}

\newtheorem*{remark}{Remark}

\theoremstyle{definition}
\newtheorem{definition}{Definition}[section]

\begin{document}

\twocolumn[
\icmltitle{The Theory and Algorithm of Ergodic Inference}




\begin{icmlauthorlist}
\icmlauthor{Yichuan Zhang}{to}
\end{icmlauthorlist}

\icmlaffiliation{to}{Department of Engineering, University of Cambridge, Cambridge, United Kingdom}

\icmlcorrespondingauthor{Yichuan Zhang}{yichuan.zhang@eng.cam.ac.uk}

\icmlkeywords{}

\vskip 0.3in
]



\printAffiliationsAndNotice{}  

\begin{abstract}
Approximate inference algorithm is one of the fundamental research fields in machine learning.
The two dominant theoretical inference frameworks in machine learning are variational inference  (VI) and Markov chain Monte Carlo (MCMC). However, because of the fundamental limitation in the theory, it is very challenging to improve existing VI and MCMC methods on both the computational scalability and statistical efficiency. To overcome this obstacle, we propose a new theoretical inference framework called ergodic Inference based on the fundamental property of ergodic transformations. The key contribution of this work is to establish the theoretical foundation of ergodic inference for the development of practical algorithms in future work.
\end{abstract}

\section{Introduction}
\label{sec:intro}
Statistical inference is the cornerstone of probabilistic modelling in machine learning. The research on inference algorithms always attracts a great attention in the research community, because it is the fundamentally important in the computation of Bayesian inference, deep generative models. The majority of research is focused on algorithmic development in two theoretical frameworks: variational inference (VI) and Markov chain Monte Carlo (MCMC). These two methods are significantly different. VI is an optimisation-based approach, in particular, which fits a simple distribution to a given target. In contrast, MCMC is a simulation-based approach, which sequentially generates asymptotically unbiased samples of arbitrary target.

Unfortunately, both VI and MCMC suffer from fundamental limitations.  VI methods are in general biased because the density function of approximate distribution must be in closed-form. MCMC methods are also biased in practice because the Markov property limits the sample simulation in a local sample space close to previous samples. However, VI is in general more scalable in computation. Optimising variational distribution and simulating samples in VI are computationally efficient and can be accelerated by parallelization on GPU. In contrast, simulating Markov chains is computationally inefficient and, more importantly, asynchronized parallel simulation of multiple Markov chains has no effect on reducing sample correlations but multiplies the computation.

Ergodic Measure preserving flow (EMPF), introduced by \citep{DBLP:journals/corr/abs-1805-10377}, is a recent novel optimisation-based inference method that overcomes the limitations of both MCMC and VI. However, there is no theoretical proof of the validity of EMPF. In this work,
we will generalize EMPF to a novel inference framework called ergodic inference. In particular, the purpose of this work is to establish the theoretical foundation of ergodic inference. 
We list the key contribution of this work as following
\begin{itemize}
\item The mathematical foundation of ergodic inference. (Section \ref{sec:ei_principle} and \ref{sec:ergodic_transformation})
\item A tractable loss of ergodic inference and the proof of the validity of the loss. (Section \ref{sec:ergodic_loss})
\item An ergodic inference model: deep ergodic inference networks (Section \ref{sec:deins})
\item Clarification of differences between ergodic inference, MCMC and VI (Section \ref{sec:deins})
\end{itemize}
\section{The background}
\label{sec:background}
Convergence of probability measures is the foundation of statistical inference. Distance metric between probability measures is critical in the study of convergence. We will review the basics of distance metrics between probability measures and connect these metrics to theoretical foundation of inference methods.

\subsection{Distance Metric of Probability Measures}
Total variation distance is fundamentally important in probability theory, because it defines the strongest convergence of probability measure.
Let $(\Omega, \cF)$ be a  measure space, where $\Omega$ denotes the sample space and $\cF$ denotes the collection of measurable subsets of $\Omega$. Given two probability measure $P$ and $Q$ defined on $(\Omega, \cF)$, the TV distance between $Q$ and $P$ is defined as
\begin{align}
\dtv(Q, P) = \sup_{A \in \cF} \vert Q(A) - P(A) \vert.
\end{align}
Convergence in TV, that is $\dtv(Q, P)=0$, means $Q$ and $P$ cannot be distinguished on any measurable set.

The Kullback-Leibler (KL) divergence is an important measure of difference between probability measures in statistical methods. For a continuous sample space $\Omega$, the KL divergence is defined as
\begin{align}
\dkl(Q \vert \vert P) = \int_{\Omega} dQ\log \frac{dQ}{dP},
\end{align}
where $dP$ denote the density of probability measure.

\subsection{Approximate Monte Carlo Inference}
Monte Carlo method is the most popular simulation based inference technique in probabilistic modelling.
For example, to fit a probabilistic model $\tar$ by maximum likelihood estimation, it is essential to compute the gradient of the partition function $Z(\theta) = \int \utar(z)dz$. Given the unnormalised density function $\log\utar(z)$, computing the gradient becomes a problem of expectation estimation
\[\partial_{\theta}Z(\theta) = \Exp_{\tar(z)}[\partial_{\theta}\log\utar(z)].\]
Monte Carlo methods allow us to construct unbiased estimator of expectation as
\[\Exp_{\tar(z)}[f(z)] = \lim_{N \rightarrow \infty}\frac{1}{N} \sum_{i=1}^N f(z_i),\]
where $z_i$ denotes samples from $\tar$.
Unfortunately, it is intractable to generate samples from complex distributions, like the posterior distributions in model parameters or latent variables.
Because of this challenge, approximate Monte Carlo Inference is fundamentally important.
We will review the theoretical foundation of two important inference methods: variational Inference (VI) and Markov chain Monte Carlo (MCMC) in the next two sections.

\subsection{Variational Inference} \label{sec:vi}
The theoretic foundation of VI is Pinsker's inequality. Pinsker's inequality states that the KL divergence is a upper bound of TV distance
\begin{align}
\dtv(Q, P) \le \dkl(Q \vert \vert P).
\end{align}
Given a parametric distribution $Q$ and the target distribution $\tar$, minimising the KL divergence $\dkl(Q \vert \vert \tar)$  implies the less TV distance $\dtv(Q, \tar)$. The key challenge of VI is how to construct the parametric family $\cQ$ so that the estimation of the KL divergence is tractable and family $\cQ$ is expressive to approximate complex target. This forces most VI methods to choose $Q$ with closed-form density function. Otherwise, the estimation of entropy term $\text{H}(Q)=-\int Q(dz) \log q(z)$ becomes challenging. In practice, the approximation family $\cQ$ in most VI methods are rather simple, like Gaussian distribution, so the approximation bias due to oversimplified $Q$ is the key issue of VI.

However, simple approximate family gives VI methods great computational advantage in practice. First, the main loss function in VI is known as the evidence lower bound (ELBO)
\begin{align}
L_{\text{ELBO}} = \int_{\Omega} dQ\log \frac{d\utar}{dQ} \le \log \int d\utar.
\end{align}
With analytic form of the entropy of $Q$, ELBO can be efficiently computed and optimized using standard gradient descent algorithm. Second, simulating i.i.d. samples from a simple variational family $Q$ is straightforward and very efficient.

\subsection{Markov Chain Monte Carlo}\label{sec:mcmc}
The theoretical foundation of Markov chain Monte Carlo (MCMC) is ergodic theorem.
Ergodic theorem states that, given an ergodic Markov chain $(Z_n)$ with a stationary distribution $\tar$, the average cross states of chain is equivalent to the average in state space of the chain, that is
\[\Exp_{\tar}[f] = \lim_{m \rightarrow \infty}\frac{1}{M} f(Z_{\infty}^{m}) = \lim_{n \rightarrow \infty}\frac{1}{N} f(Z_n),\]
where $Z_{\infty}^m$ denotes the sample of a well-mixed Markov chains after infinitely long transitions.
Ergodic theorem implies that we can generate unbiased samples from every Markov transition without waiting forever for the chains to reach stationary state. Therefore, we can trade computational efficiency with a bias that may decrease in a long time.
The key challenge of MCMC methods is to define ergodic Markov chains with any given stationary distribution $\tar$. This challenge was solved first by Metropolis-Hastings algorithm. We will discuss in detail in Section \ref{sec:mh}.

Ergodic Markov chains enjoy strong stability. Irrespective of the distribution of initial state $\mu(z_0)$ and the parameter of Markov kernel $K(\cdot, \cdot)$, the distribution of the state of the chain is guaranteed to converge to the stationary distribution in total variation after every transition. Formally, that means the reduce of TV distance to stationary for all $L \ge 0$
\begin{align*}
\dtv\left(Q_{L+1}, \tar\right) < \dtv\left(Q_{L}, \tar\right)
\end{align*}
where $q_{L}$ denotes the marginal distribution of the $L$-th state and
\begin{align*}
q_{L}(dz') = \int K(z, d z')q_{L}(dz).
\end{align*}
As $L$ increases, the distribution $q_L$ converges to a unique stationary distribution $\tar$
\[\lim_{l \rightarrow \infty}\dtv(Q_l, \tar) =0.\]

In spite of the theoretical convergence property, the convergence of MCMC chains is not guaranteed in practice. Because the burn-in stage cannot be infinite long, the samples from MCMC methods are often biased. The problem is that there is no reliable measurement of such a sampling bias related to TV distance or KL divergence. The iterative simulation of Markov chain is another limitation in computational efficiency. Each sample from MCMC methods requires one simulation of Markov transition and this can only be executed in a sequential manner due to the nature of Markov chain. Therefore, the sampling time of MCMC grows linearly with the number of samples.

\section{Ergodic Inference Principle}
\label{sec:ei_principle}
In this section, we present the mathematical foundation of ergodic inference principle.

\subsection{Motivation}
First, we would like to propose the the following properties of ideal inference method:
\begin{itemize}
\item Parallelizable: the simulation of each sample is computationally independent;
\item Statistically efficient: there is zero correlation between samples;
\item Asymptotic unbiased: more computational power guarantees diminishing of simulation bias. The bias can be eliminated in theory with sufficient computation.
\end{itemize}
Both MCMC and VI fail to have all the properties above. For this reason, there are existing works on a hybrid methods that combine MCMC and VI, for example, accelerate the burn-in of MCMC using variational approximation in \cite{pmlr-v70-hoffman17a} or optimise ELBO based on tractable density function of MCMC kernel in \cite{salimans2015markov}. To some extend, such algorithmic hybrid approach can be useful in practice. However, the limitation in theoretical foundation of MCMC and VI cannot be eliminated by algorithmic modification. To achieve an ideal inference method, it is necessary to have a new mathematical theoretical foundation.

\subsection{The Theoretical Foundation}
\label{sec:foundation}
Different from Pinsker's inequality and ergodic theorem,
the theoretical motivation of the proposed inference is the characteristic property of ergodic Markov transition: \emph{there is a unique invariant distribution for every ergodic Markov Kernel}.
Formally, let $K_{\tar}$ be an ergodic Markov transition kernel with an invariant distribution $\tar$. By construction of $K_{\tar}$, $\tar$ is guaranteed to be the only distribution satisfies the condition $\tar(d\bz') = \int K_{\tar}(\bz, d\bz') \tar(d\bz)$.

Based on the property of ergodic Markov kernel, we construct the following criteria to verify if a distribution is equivalent to the stationary distribution of the kernel.
Given a distribution $q$, the distribution of $q$ after one Markov transition by $K$ is given by
\begin{align}
q_1(\bz')  = \int K_{\pi}(\bz, \bz') q(d\bz).
\label{eq:q_1}
\end{align}
We say the distribution $q$ is preserved by $K_{\pi}$ if 
\begin{align}
\dtv(q_1, q) = 0.
\label{eq:db_L2}
\end{align}
By the uniqueness of the invariant distribution of ergodic kernel $K_{\pi}$, the preservation of $q$ by $K_{\pi}$ as \eqref{eq:db_L2} implies $\dtv(q, \tar) = 0$.
This motivates the following loss function.
\begin{definition}
Given a Markov kernel $K_\tar(\bz, \bz')$ that is ergodic w.r.t. a distribution $\tar$, the ergodic loss of a distribution $q$ is defined as 
\[L^*(q, K_{\tar}) = \dtv\left(\int K_{\tar}(\bz, \cdot)q(d\bz), q(\cdot) \right). \]
\end{definition}
As mentioned earlier, the loss $L^*(q, K_{\tar}) $ is equal to 0 if and only if $\dtv(q, \tar)$ is equal to 0.

Let $\tar$ be the target distribution and $q$ be the approximate distribution in a parametric family $\cQ$. Given an ergodic Markov kernel $K_{\pi}$, the closest $q \in \cQ$ to the target $\tar$ can be identified by the parameter $\phi^*$ optimising the ergodic loss $L^*(\cdot, K_{\pi})$
\[\phi^* = \argmin_{\phi}L^*(q_{\phi}, K_{\pi}).\]
If the target distribution is in $\cQ$, then the optimal parameter $\phi^*$ should have the loss
\[L^*(q_{\phi^*}, K_{\tar}) =0,\]
otherwise the $L^2$ norm of the gradient of the loss should vanish
\[\vert\vert \partial_{\phi^*} L^*(q_{\phi^*}, K_{\tar}) \vert\vert_{2}^2 = 0.\]
\subsection{Technical Challenges}
There are two technical challenges of ergodic inference methods in practice. First, we need a tractable estimation of a loss function equivalent to $\dtv(q_1, q)$. The estimation of the gradient of the loss should also be tractable for the optimisation of the parameter $\phi$. Second, we need a general parametric family $\cQ$ that can approximate any target distribution up to a certain amount of error. More specific, the error can be controlled and even eliminated by increase the complexity of approximation family of $\cQ$, i.e. the number of parameters of $\cQ$ is unlimited.
The computational cost of optimisation is associated with the complexity of $\cQ$.

We will present the solution to the first challenge in Section \ref{sec:ergodic_loss} and the solution to the second challenge in Section \ref{sec:deins}.

\section{Ergodic Transformations}
\label{sec:ergodic_transformation}
The key of solving the technical challenges in ergodic inference is the reparameterization of the ergodic Markov kernel. This is important in both algorithmic development and theoretical analysis.
\subsection{Ergodic Transformations and Markov Kernels}
Ergodic Markov kernels are essentially conditional distributions, which can be reparameterized by deterministic transformations known as measure preserving transformations (MPTs).
Given a probability measure $\mu$, a deterministic transformation $T$ preserves $\mu$ if for any measurable subset of sample space $A$, $\mu(T^{-1}(A)) = \mu(A)$.
The shear transformation $T(x, y)=(x+y, x)$, which preserves Lebesgue measure, is a classic example of MPT \cite{Bill86}.
The following conditions are often used in the literature MCMC theory for verification of ergodic property:
\begin{enumerate}
\item Irreducibility: $T(A) \ne A\,, \forall A \in \cF$ except $\emptyset$ and $\Omega$.
\item Density preservation: $\tar(T(\bz)) = \tar(\bz)$.
\item Lebesgue preservation: the determinant of the Jacobian of $T$ is equal to 1.
\end{enumerate}
Formally, we define the reparameterisation of Ergodic Markov chains as following.
\begin{definition}
\label{def:ergodic_reparam}
(Ergodic Reparameterisation of MCMC)
Given a target distribution $\tar(\bz)$, a MCMC kernel $K(\bz, \bz')$ with invariant $\tar$ can be reformed as two steps:
\begin{enumerate}
\item Simulate an auxiliary variable $\br$ with distribution $\mu(\br)$
\item Deterministic transformation $(\bz', \br') = T_{\tar\mu}(\bz, \br)$,
\end{enumerate}
where $T_{\tar\mu}$ is an ergodic transformation that preserves the probability measure $\tar(\bz)\mu(\br)$.
\end{definition}

\begin{remark}
The transformation $T_{\tar\mu}$ in ergodic reparameterisation is fundamentally different from volume preserving transformation $V(\bz)$ in the sample space of $\bz$ for two reasons.
\begin{itemize}
\item $T_{\tar\mu}(\bz, \br)$ does not preserve the volume/entropy in the sample space of $\bz$, but $V(\bz)$ must preserves the volume/entropy in the space of $\bz$.

\item $T_{\tar\mu}(\bz, \br)$ preserves the probability measure $\tar(\bz)$, but $V(\bz)$ does not preserve $\tar(\bz)$ in general.
\end{itemize}
\end{remark}
Ergodic transformations also allow us to form the expectation under Markov transition as composition of functions, that is not used in classic MCMC literature. Formally, this is given by the following proposition.  
\begin{proposition}
\label{proposition:mpt}
Given an ergodic transformation $T_{\pi}$ w.r.t. $\pi$, the expectation is preserved by the transformation, which means, for any function $f$
\[\int_{\Omega} f(\bz) \pi(d\bz) = \int_{\Omega} f \circ T_{\pi}(\bz) \pi(d\bz) = \int_{\Omega'} f(\bz') T_{\pi *}\pi(d\bz'),\]
where $\Omega'$ is the image of $\Omega$ under $T_{\pi}$ and $T_{\pi *}\pi(\cdot)$ denotes the pushforward probability measure of $\pi$ under $T_{\pi}$. Because $T_{\pi}$ preserves $\tar$, $\Omega' = \Omega$ and $\dtv(\pi, T_{\pi *}\pi) = 0$.
\end{proposition}

In the next two sections, we will demonstrate the ergodic reparameterization with two well-known MCMC kernels.

\subsection{Metropolis-Hastings Transformations}
\label{sec:mh}
Metropolis-Hastings (MH) algorithm is the first and most well-known MCMC methods. We will show that it is straightforward to form the MH transition kernel as an ergodic transformation.
Given a target distribution $\pi(\bz)$ and a transition proposal distribution $q(\br \vert \bz)$,  MH kernel in most text books is described as following two steps:
\begin{enumerate}
\item Sample $\br$ from $q(\cdot \vert \bz)$. 
\item Return the new state of the chain as $\br$ with probability
\begin{align}
p_{MH} = \min\left\{1, \frac{\pi(\br)q(\bz\, \vert\, \br)}{ \pi(\bz)q(\br\, \vert\, \bz)}\right\},
\end{align}
otherwise the state remains as $\bz$.
\end{enumerate}

It is straightforward to verify that MH transition kernel preserves the density function as
\begin{align*}
&\pi(\bz)\left[q(\br \vert \bz)\min\left\{1, \frac{\pi(\br)q(\bz\, \vert\, \br)}{ \pi(\bz)q(\br\, \vert\, \bz)}\right\}\right]\\
=&\min\left\{\pi(\bz)q(\br\, \vert\, \bz), \pi(\br)q(\bz\, \vert\, \br)\right\}\\
=&\pi(\br)\left[ q(\bz \vert \br)\min\left\{1, \frac{ \pi(\bz)q(\br\, \vert\, \bz)}{\pi(\br)q(\bz\, \vert\, \br)} \right\} \right],
\end{align*}
where the MH transition kernel $K_{MH}(\cdot, \cdot)$ is in squared rackets.
This verification of stationary distribution is known as detailed balance. It is important because it  proves the existence of stationary distribution. 

Now we consider an alternative representation of MH kernel. In particular, we define a stationary distribution as the joint distribution of all random variables involved in the target $\pi$ and MH kernel $K_{MH}$, that is $\pi(\bz, \br, u) = \pi(\bz) q(\br \vert \bz)\nu(u)$, where $\nu(u)$ denotes uniform distribution between $[0, 1]$.
Following the ergodic reparameterization (Definition \ref{def:ergodic_reparam}), we can rewrite the MH algorithm as
\begin{enumerate}
\item Resample $\br$ from $q(\cdot \vert \bz)$ and $u$ from $\nu(\cdot)$.
\item Return the next state $(\bz', \br', u') = T_{MH}(\bz, \br, u)$ defined as
\begin{align}
T_{MH}(\bz, \br, u) &= (\bz, \br, u)\delta(u>p_{MH}) \nonumber \\
&+ (\br, \bz, u)\delta(u<p_{MH}),
\label{eq:mh_func}
\end{align}
where $\delta(\cdot)$ denotes indicator function.
\end{enumerate}
Notice that the transformation $T_{MH}(\bz, \br, u)$ above is a deterministic function.
It is obvious that resampling $\br$ and $u$ from their conditional distribution leaves $\pi(\bz, \br, u)$ invariant. Then, it is straightforward to show the preservation of density function
\[\pi(\bs)\delta(\bs' = T_{MH}(\bs)) = \pi(\bs')\delta(\bs = T_{MH}(\bs')),\]
where $\bs$ denote the triple $(\bz, \br, u)$.
It is also easy to verify that the determinate of Jacobian of $\partial_{(\bz, \br, u)}T_{MH}(\bz, \br, u)$ is always equal to 1. 

\subsection{Hamiltonian Measure Preserving Transformations}
Hamiltonian Monte Carlo (HMC), originally known as Hybrid Monte Carlo, is an important MCMC method. Originally, HMC is considered as a hybrid method, because its combines both deterministic and stochastic simulation.
The deterministic simulation in HMC essentially refers to any dynamics that generalize the classic Hamiltonian dynamics in physics.

Hamiltonian system in physics is a system of moving particles in an energy field and the energy of the system is constant over time. Given $n$ particles, the state of Hamiltonian system is defined by the position  $\bz \in \Real^n$ and the momenta $\br \in \Real^n$. The position $\bz$ is associated with potential energy $U:\Real^n \rightarrow \Real$ and the momentum $\br$ is associated with kinetic energy $K:\Real^n \rightarrow \Real$.
The state $(\bz, \br)$ evolves over time $t$, according to Hamilton’s equations:
\begin{align}
\dot \bz(t) = \partial_{\br}K(\br);\, \dot \br(t) = -\partial_{\br}U(\bz),
\end{align}
where $\dot \bz$ denotes the derivative of $\bz$ w.r.t. time $t$.
It is straightforward to verify that the total energy $H = U + K$ does not change over time
\[\dot H(\bz, \br) = \left(\partial_{\br}U(\bz)\right)^T\partial_{\br}K(\br) -  \left(\partial_{\br}U(\bz)\right)^T\partial_{\br}K(\br) = 0.\]
Given an initial condition $(\bz, \br)$, the solution of Hamiltonian dynamics is a function of time $t$
\[(\bz(t), \br(t)) = T_{H}(t, \bz, \br).\]
Given a fixed time $t$, the solution $T_{H}$ becomes a map $T_{H, t}: \Real^{2n} \rightarrow \Real^{2n}$ between two states $(\bz, \br)$ and $(\bz', \br')$ with the same total energy $H$.
Intuitively, $\bz(t)$ forms a trajectory of particle traversing in a $n$-dimensional space and the velocity of the particle is given by $\dot \bz(t) = \partial_{\br}K(\br(t))$.

It is well-known in MCMC literature that $T_{H, t}$ is essentially a family of measure preserving transformations with any parameter $t \in \Real \ne 0$. It is clear that $T_{H, t}$ is irreducible if $t \ne 0$ and density preserving w.r.t. $\exp(-H)$. The volume preservation property of Hamiltonian dynamics in the state space $(\bz, \br)$ is a well-known result of Liouville’s Theorem. Therefore, we know that $T_{H, t}(\bz, \br)$ with any $t \ne 0 $ is an ergodic transformation w.r.t. the distribution $\pi(\bz)\mu(\br) \propto \exp(-H(\bz, \br))$. This implies $T_{H, t}$ also preserves $\pi \propto \exp(-U)$ by the definition of marginal distribution.

In practice, Hamiltonian dynamics do not have closed-form solutions. Fortunately, there is a rich literature on the numeric simulation of Hamiltonian dynamics. The most known approximate approach in HMC is Leapfrog algorithm, which is constructed as a sequential of shear transformations. Leapfrog algorithm enjoys strong stability and good approximation error is around squared discretized step size. See more detailed analysis in \cite{radford2010, leimkuhler2004simulating}.
\section{Ergodic Loss}
\label{sec:ergodic_loss}

\subsection{$\tar$-Ergodic Loss Function}

By the definition of TV distance, we know that $q$ is the stationary distribution of $K$ if and only if for all function $f(\cdot)$ with $\Exp_{\pi}[f(\bz)] < \infty$,
\begin{align}
\Exp_{q_1}[ f(\bz) ] = \Exp_{q}[ f(\bz)].
\label{eq:db_L}
\end{align}
However, it is impossible to compare the expectation of all possible function $f$, but given specific function $f$ it is possible to estimate
\begin{align}
L_{K, f}(\phi) = \vert \Exp_{q_1}[f(\bz)] - \Exp_{q}[ f(\bz)] \vert \,.
\label{eq:f-ergodic_Loss}
\end{align}
With the optimal choice of function $f$ and certain condition, we can claim that $L_{K, f}(\phi) = 0$ implies $\dtv(q, \tar) = 0$. The log density function is an intuitive choice, because we can identify a distribution by its density function. Therefore, we define the following $\tar$-ergodic loss.
\begin{definition}(Ergodic Loss Function)
\begin{align}
L_{K, \tar}(\phi) = \vert \Exp_{q_1}[ \log \tar(\bz) ] - \Exp_{q}[ \log \tar(\bz) ] \vert\,.
\label{eq:db_Loss_1_LL}
\end{align}
\end{definition}


\begin{theorem} (Ergodic Loss Convergence Theorem)\label{theorem:loss_convergence}
Given the ergodic Markov kernel $K_{\tar}$ with invariant distribution $\tar$, the loss $L_{K, \tar}(\phi) = 0$ if and only if $\Exp_{\tar}[ \log \pi(\bz) ] = \Exp_{q}[ \log \pi(\bz) ] $.
\end{theorem}
\begin{proof}
The convergence of loss $L_{K, \tar}(\phi)=0$ implies
\begin{align}
 \Exp_{q_1(\bz)\mu(\br)}[ \log \tar(\bz) ] = \Exp_{q(\bz)\mu(\br)}[ \log \tar(\bz) ]\,,
 \label{eq:theorem1_1}
\end{align}
where $q_1(\bz)$ is given by \eqref{eq:q_1}. Notice that $q_1$ is essentially the marginal of the pushforward of $q(\bz)\mu(\br)$ under the measure preserving transformation $T_{\tar\mu}$. By Proposition \ref{proposition:mpt}, the expectations in \eqref{eq:theorem1_1} can be written as following 
\begin{align}
\Exp_{q_1(\bz)}[ \log \tar(\bz) ] \overset{\Delta}{=} \int_{\Omega} \log\tar \circ T_{\tar\mu}\, d(q\mu) = \int_{\Omega} \log\tar\, d(q\mu),
 \label{eq:theorem1_10}
\end{align}
where $d(q\mu)$ is the shorthand notations for $q(\bz)\mu(\br)d\bz d\br$.
Replacing $q\mu$ on both sides in \eqref{eq:theorem1_10} with any distribution, the equality still holds. If we replace $q\mu$ in \eqref{eq:theorem1_10} with with the pushforward probability measure of $q(\bz)\mu(\br)$ under $T_{\tar\mu}$, denoted by $T_{\tar\mu*} (q\mu)$, we have
\begin{align*}
\int_{\Omega} \log\tar \circ T_{\tar\mu} \circ d(T_{\tar\mu*}(q\mu)) = \int_{\Omega} \log\tar\, \circ d(T_{\tar\mu*}(q\mu)),
\end{align*}
which can be rewritten as
\begin{align}
\int_{\Omega} \log\tar \circ T_{\tar\mu}^1\circ T_{\tar\mu}\, d(q\mu^1)\, = \int_{\Omega} \log\tar\circ T_{\tar\mu}\, d(q\mu),
 \label{eq:theorem1_11}
\end{align}
where $T_{\tar\mu}^1$ denotes $T_{\tar\mu} = (\bz, \br_1)$ and $d\mu^1$ denotes $\mu(d\br_1)$.
Notice that the LHS of \eqref{eq:theorem1_11} is an expectation under the distribution of $\bz$ after two ergodic Markov transitions from $q$, that is $\Exp_{q_2(\bz)}[ \log \tar(\bz) ]$.
Therefore, by \eqref{eq:theorem1_10} and \eqref{eq:theorem1_11}, we have
\begin{align}
\Exp_{q_2(\bz)}[ \log \tar(\bz) ] &\overset{\Delta}{=} \int_{\Omega} \log\tar \circ T_{\tar\mu}^1\circ T_{\tar\mu}\, d(q\mu^1)\, \nonumber \\
&= \int_{\Omega} \log\tar\circ T_{\tar\mu}\, d(q\mu) \nonumber \\
&= \Exp_{q(\bz)}[ \log \tar(\bz) ].
 \label{eq:theorem1_12}
\end{align}
By induction, we know the expectation of $\Exp_{q_n}[\log \tar]$ does not change after any number of measure preserving transformation $T_{\tar\mu}$, that gives
\begin{align}
\Exp_{q_{\infty}(\bz)}[ \log \tar(\bz) ] = \Exp_{q(\bz)}[ \log \tar(\bz) ].
 \label{eq:theorem1_13}
\end{align}
By \eqref{eq:theorem1_13}, we know if we simulate infinitely long ergodic Markov chain by kernel $K_{\pi}$, then the expectation $\Exp_{q_{\infty}(\bz)}[ \log \tar(\bz) ]$ is the same as the initial expectation $\Exp_{q(\bz)}[ \log \tar(\bz) ]$.

Because an ergodic Markov chain has unique invariant distribution, \eqref{eq:theorem1_13} implies
\begin{align}
\Exp_{\tar(\bz)}[ \log \tar(\bz) ] = \Exp_{q(\bz)}[ \log \tar(\bz) ].
\label{eq:theorem1_14}
\end{align}
\end{proof}

Recall that the convergence of loss $L_{K, \utar}(\phi)$ cannot be sufficient for the convergence of the TV distance $\dtv(q, \tar)=0$. Fortunately, under some reasonable condition, the loss $L_{K, \utar}(\phi) = 0$ implies the convergence in TV distance. Formally, this is given by the following theorem.
\begin{theorem} (Ergodic Measure Convergence Theorem)\label{theorem:ergodic_fundamental_theorem}
Let $K_{\pi}$ be an ergodic Markov kernel with invariant distribution $\tar$. Assume that the entropy of $Q$ is not less than the entropy of $\tar$, that is $\text{H}(Q) \ge \text{H}(\tar)$, the loss $L_{K, \utar}(\phi) = 0$ if and only if $\dtv(q, \tar) = 0$.
\end{theorem}
\begin{proof}
By the definition of the KL divergence, we have
\begin{align}
\dkl(q \vert \vert \tar) = \Exp_q[\log q] - \Exp_q[\log \tar].
\end{align}
By Theorem \ref{theorem:loss_convergence}, we have
\begin{align}
\dkl(q \vert \vert \tar) = \Exp_q[\log q] - \Exp_\tar[\log \tar],
\end{align}
which is equivalent to
\[\dkl(q(\bz) \vert \vert \tar) = H(\tar) - H(Q).\]
Because the KL divergence is never less than 0, we have
\[H(\tar) \ge H(Q).\]
Finally, by the assumption $H(\tar) \le H(Q)$, we know $H(\tar) = H(Q)$,
so we know
$0\le \dtv(q, \tar) \le \dkl(q \vert \vert \tar) = 0,$
which implies $\dtv(q, \tar) = 0$.
\end{proof}

By the monotonic convergence in TV distance of ergodic Markov chain, it is straightforward to show that
\begin{proposition}\label{prop:loss}
Given a smooth ergodic transformations w.r.t. the probability measure $\tar(\bz)$, if $\Exp_{q}[\log\tar] < \Exp_{\tar}[\log\tar ]\,$, the loss 
\begin{align}
\Exp_{q}[\log \utar(\bz)] - \Exp_{q_1}[ \log \utar(\bz) ] > 0.
\label{eq:final_objective2}
\end{align}
\end{proposition}
Assume that $\Exp_{q}[\log\tar] < \Exp_{\tar}[\log\tar ]\,$, we have
\begin{align}
L_{K, \utar}^*(\phi) = \Exp_{q}[\vert \log \pi^*(\bz) \vert ] - \Exp_{q_1}[\vert \log \pi^*(\bz) \vert ] \,,
\label{eq:db_Loss_2_LL}
\end{align}
\subsection{Optimising $\utar$-Ergodic Loss}
\label{sec:optimising_ergodic_loss}

Let $q_{01}(\bz, \bz_1)$ be the joint distribution $q(\bz)K(\bz, \bz_1)$. Then, we can rewrite \eqref{eq:db_Loss_2_LL} as
\begin{align}
L_{K, \utar}^*(\phi) = \Exp_{q_{01}}[\log \pi^*(\bz)  - \log \pi^*(\bz_1) ] \,,
\label{eq:db_Loss_3_LL}
\end{align}
which can be estimated by samples of $(\bz, \bz_1)$.
To optimise the loss \eqref{eq:db_Loss_3_LL}, we need to compute the gradient $\partial_{\phi}L_{K, \utar}^*(\phi)$.
Notice that the $\bz$ and $\bz_1$ are coupled by the kernel $K$ and the density function of most MCMC kernels, which makes the computation of the gradient $\partial_{\phi}L_{K, \utar}^*(\phi)$ unstable. To avoid this, we reparameterize both $q(\cdot)$ and the ergodic Markov kernel $K(\bz, \cdot)$ by a transformation $T_{\phi}$ and a measure preserving transformation $T_{\tar}$ respectively. This allows us to transform some simple random variable $\br$ and $\br_1$, that is independent of $\phi$, into $(\bz, \bz_1)$ as
\begin{align}
\bz = T_{\phi}(\br),\quad
\bz_1 = T_{\tar}(\bz, \br_1).
\label{eq:db_Loss_3_LL_part_2}
\end{align}
Therefore, we can compute the loss with following reformulation
\begin{align}
L_{K, \utar}^*(\phi) = \Exp_{\mu(\br)\mu_1(\br_1)}[L_{\utar, T_{\phi}, T_{\tar}}(\br, \br_1) ] \,,
\label{eq:db_Loss_3_LL}
\end{align}
where $L_{\utar, T_{\phi}, T_{\tar}} = \log \utar(\bz) - \log \utar(\bz_1)$ and $(\bz, \bz_1) = T_{\phi, \pi}$ as \eqref{eq:db_Loss_3_LL_part_2}.

As discussed above, the only requirement of approximate family $\cQ$ in ergodic inference is the transformation $T_{\phi}$ is known and it is a measurable function. It is an important advantage over VI, where the density function of $\cQ$ must be in closed form.


\section{Deep Ergodic Inference Model}
\label{sec:deins}
Ergodic transformations are not only fundamentally important in the ergodic loss, they are also powerful tools for constructing flexible approximation family $\cQ$. In this section, we will present how to construct and optimise the approximation family $\cQ$ by stacking multiple layers of ergodic transformations. 
\subsection{Definition}
Let $\{K_1, K_2, \dots, K_N\}$ be $N$ ergodic transition kernel with independent parameters $\{\phi_1, \phi_2, \dots, \phi_N\}$. Let $q_0$ be the distribution of initial state also has parameter $\phi_0$.
By ergodic reparameterization, we reform each ergodic Markov kernel $K_n(\bz, \bz')$ as a transformation $\bz_{n} = T_{n}(\bz_{n-1}, \br)$, where $T_n$ is a deterministic function depends on the kernel parameter  $\phi_n$ and $\br$ is sampled from a standard distribution $\mu_n$. We also reparameterize the initial distribution $q$ from a simple distribution $\mu_0$ by a transformation $T_0$. Then, we can generate samples of $\bz_n$ by transforming samples of $(\br_0, \br_1,\dots,\br_{N-1})$ from $\mu(\cdot) = \prod_{i=0}^{N-1} \mu_i(\cdot)$ as
\begin{align}
\bz_n = T_{\br_{N-1}} \circ \cdots \circ T_{\br_{1}} \circ T_0(\br_0),
\label{eq:dein_def}
\end{align}
where $T_{\br_{n}}(\cdot)$ denotes $T_{n}(\cdot, \br_n)$. 
We call this multiple layer ergodic transformation $T_{\br_N-1} \circ T_{\br_1} \circ T_0(\cdot)$ deep ergodic inference network (DEIN). The expectation of $q_N$ can be reformed as
\[\Exp_{q_N}[f(\bz_N)] = \Exp_{\mu}[f\circ T_{\br_{N-1}} \circ \cdots \circ T_{\br_{1}} \circ T_0(\br_0)],\]
which allows us to estimate the gradient of any function by Monte Carlo method
\[\partial_{\bphi} \Exp_{q_N}[f(\bz_N)] \approx \frac{1}{M} \sum_{i=1}^M\partial_{\bphi} f\circ T_{\br_{N-1}^i} \circ \cdots \circ T_{\br_{1}^i} \circ T_0(\br_0^{i})\,.\]

\subsection{Optimisation and Convergence of DEINs}
This is a non-parametric model because the number of parameters of this model grows with the number of transformations. Different from deep neural networks, DEIN has strong stability by the natural of ergodicity. In particular, DEINs can be arbitrarily deep and the stability and simulation quality is guaranteed to improve with the depth.

First, we define a loss \eqref{eq:db_Loss_1_LL} for each transition $K_n$ as
\begin{align*}
L^n(\phi_n) = \Exp_{q_n}[\log \pi^*(\bz)] - \Exp_{q_{n-1}}[\log \pi^*(\bz)] \,,
\end{align*}
where $q_N$ denotes the marginal of the last state
\begin{align}
q_n(\bz; \bphi_{0:n}) = \int  K(\bz_{n-1}, \bz_n) q_{n-1}(\bz_{n-1}; \bphi_{0:n-1}).
\label{eq:Dein_density}
\end{align}
\begin{proposition}\label{prop:loss_dein}
Assume that $ \Exp_{q_0}[\log \pi(\bz)] <  \Exp_{\tar}[\log \pi(\bz)]$, minimizing the ergodic loss $L^*_{K, \utar}$ in \eqref{eq:db_Loss_2_LL} with $q_N$ of deep ergodic Inference network is equivalent to maximizing the total ergodic loss $\sum_{n=1}^N L^n(\phi_n)$
\begin{align}
L_N(\bphi) = \Exp_{q_N}[\log \pi^*(\bz)] - \Exp_{q_0}[\log \pi^*(\bz)]\,.
\label{eq:Loss_all}
\end{align}
which is equivalent to
\begin{align}
L_N(\bphi; \phi_0) = \Exp_{q_N}[\log \pi^*(\bz)].
\label{eq:Loss_empf}
\end{align}
when the parameter of $q_0$ is fixed.
\end{proposition}
The total loss \eqref{eq:Loss_empf} is consistent with the loss proposed by \cite{DBLP:journals/corr/abs-1805-10377} in ergodic measure preserving flows.

By Proposition \ref{prop:loss}, it is straightforward to show that DEINs enjoy incremental improvement as the depth grows.
\begin{theorem} (Incremental Convergence of DEIN)
\label{theorem:incremental_convergence}
Given a $N$-layer DEIN defined as \eqref{eq:dein_def}, the optimal total ergodic loss $L_N(\bphi^*) = \max_{\bphi}L_N(\bphi)$ increases monotonically as $N$ increases.
\end{theorem}

Similar to the convergence of ergodic Markov chains, we have the asymptotic unbiased convergence of DEINs as following.
\begin{theorem} (Asymptotic Unbiased Convergence of DEINs)
\label{theorem:incremental_convergence}
For arbitrarily small $\epsilon>0$, there always exists a DEIN with finite number of layer $N$, so that
with the optimal distribution $q_N^*$ has the ergodic loss $L_{K, \tar}^* = \dtv\left(\int K(\bz, \cdot)dq_N^*, q_N^*(\cdot)\right) \le \epsilon$.
\end{theorem}

\subsection{Comparison with Auto-Tuning MCMC}
From an algorithmic perspective, auto-tuning MCMC (AMCMC) and DEIN are very similar, because both methods simulate ergodic Markov chains and optimise the parameters of the kernel w.r.t. a loss. This may give a false impression of that AMCMC and DEIN share the same theoretical foundation. 

To clear this impression, we will discuss the fundamental difference between DEINs and AMCMC.
First of all, AMCMC is essentially a class of MCMC methods with auto-tuning strategy of kernel parameters.
In particular, the purpose of auto-tuning is to boost the statistical power of samples from MCMC by encouraging distant jump between states in Euclidean space, which is inspired by the work of \citep{pasarica2010adaptively} on reducing sample correlation of MCMC.
In contrast, as a parametric family in ergodic inference methods. The parameters in DEINs is optimised w.r.t. the ergodic loss, which is based on the ergodic inference principle in Section \ref{sec:foundation}. 

The fundamental difference have two important effects in practice. 
The first effect is on the sample correlation.
By the nature of Markov property, optimising the auto-tuning loss can never eliminate the correlation of samples from MCMC.
In contrast, the samples from DEINs are generated by deterministic transformation of i.i.d. samples from initial distribution, which is still i.i.d. samples.
The second consequence is on the MH-correction.
In particular, MH correction is optional for DEINs for three reasons.
First, DEIN is a parametric approximate family $\cQ$ rather than unbiased simulation procedure.
Second, by optimising the ergodic loss, DEINs guarantee the convergence towards the target in TV distance. Finally, even with approximate ergodic transformations, the existence of a stationary distribution (not necessarily the target) is guaranteed by measure preserving property, in  particularly with the depth of DEIN is always finite.
In contrast, the convergence of AMCMC chains is only guaranteed with MH correction. In particular, without MH correction, the existence of a stationary distribution of MCMC chains becomes questionable. With unlimited number of recurrent Markov transitions, Markov chains are not guaranteed to converge to any distribution. The existence of stationary distribution is the necessary condition of ergodic theorem \cite{Robert:2005:MCS:1051451}. Therefore, without MH-correction (implicitly proved by detailed balance condition), the bias of samples from MCMC may not be bounded. This is particularly true when the Markov kernel parameter is tuned to maximize the jumping distance between states.

\subsection{Comparison with Normalising Flows}
Normalizing Flow (NF),  introduced by \citep{Rezende:2015:VIN:3045118.3045281}, is a recent variational inference framework, where the variational parametric distribution is defined in an iterative procedure. The fundamental idea of NF is to define an expressive parametric family by a sequence of deterministic transformations with closed-form Jacobian. Let $\bz_0$ be a random variable from a simple distribution $\mu$, like Gaussian, and $f_1\dots, f_M$ be $M$ deterministic functions from $\Real^n$ to $\Real^n$. We define a sequence of random variable ${\bz_1\dots\bz_M}$ as
\[\bz_M = f_M \circ\dots \circ f_1(\bz_0).\]
By the rule of changing variables, the density function of $\bz_M$ is given by
\[\log p(d\bz_M) = \log q(d\bz_0) - \sum_{i=1} \log \left\vert \det \partial_{\bz_i} f_i(\bz_i)\right \vert.\]

There are three important difference between DEINs and NFs.
First, without manually engineering ergodic transformations, DEINs have theoretical guarantee of better performance with more transformations (Theorem \ref{theorem:incremental_convergence}). In contrast, the transformations $f_i$ in NFs is predefined based on heuristics and experimental evidence.
Second, ergodic transformations $T_{\tar}$ has no closed form solutions, but the transformations $f_i$ in NFs is limited to simple functions with tractable Jacobian.
Finally, the distribution of DEINs is very expressive, which may not even have a closed form as \eqref{eq:Dein_density}. More importantly, there is no need to compute the density for optimising the parameters. It is the opposite for NFs. In particular, the transformations in NFs are often restricted to simple functions to have closed-form Jacobian. The computation of the Jacobian is also one of computational bottlenecks in optimisation.

\subsection{Comparison Overview}
The key difference between ergodic inference, AMCMC and VI is highlighted in the following table.
\begin{center}
\resizebox{\linewidth}{!}{
  \begin{tabular}{ | c | p{2cm}  | p{2cm} | p{2cm} |}
    \hline
    Method & TV-Loss & Implicit Simulation Density  & Independent samples \\ \hline
    VI & Yes & No & Yes\\ \hline
    AMCMC & No & Yes & No\\  \hline
    EI & Yes & Yes & Yes\\ 
    \hline
  \end{tabular}
 }
\end{center}
\begin{itemize}
\item TV-Loss: Optimising the loss function leads to the convergence in TV distance.

\item Independent samples: computationally and statistically independent sample simulation.

\item Implicit Simulation Density: no closed-form density function of simulation distribution is required in training.
\end{itemize}

\section{Related Works}
Hamiltonian variational inference (HVI), introduced by \citep{salimans2015markov}, is an interesting variational framework using MCMC kernel as variational parametric distribution.
The motivation of HVI is that the joint density function of all the states of HMC chains is tractable to compute. Unfortunately, the variational lower bound is still intractable to compute, because the reverse probability of HMC chain given the last state is intractable. To overcome this problem, they propose to approximate the reverse density function using neural network. Although HVI shows improvement in performance over VAEs, the additional approximation limits the potential of this method. However, optimising the HMC kernel parameters w.r.t. ELBO is still an attractive feature of HVI.

\citeauthornum{pmlr-v70-hoffman17a} proposed another hybrid method based on VI and HMC without auxiliary approximation. The idea is to use a Monte Carlo estimation of the marginal likelihood by averaging over samples from HMC chains, that are initialized by variational distribution. In \citep{han2017alternating} a very similar framework is proposed using Metropolis-adjusted Langevin dynamics. This idea is very similar to contrastive divergence in \cite{Hinton02}. The main disadvantage of this methods is that the HMC parameters are manually pretuned. Especially, As mentioned by \citep{pmlr-v70-hoffman17a}, No-U-turn Sampler (NUTS), an adaptive HMC, is not appliable due to engineering difficulties. \cite{radford2010} pointed out that HMC is very sensitive to the choice of Leapfrog step size and number of leaps. 

Stein Variational Gradient Descent (SVGD) is a recent particle based dynamical inference method proposed by \citep{NIPS2017_6904}.
In SVGD, the approximation distribution is a set point mass $q$ generated by transforming a set of points sampled from a distribution $\mu$ using  a perturbation function $T(x) = x + \phi(x)$, where $\phi$ is in a function space with boundary norm. With this setup, the optimisation of $T$ w.r.t. the KL divergence between $q$ and the target $\tar$ is transformed into a stochastic optimisation in the kernel space of $\phi$. The theoretical foundation of convergence of SVDG is sound and appealing. However, this method faces two practical challenges. First, the optimisation complexity grows quadratically with the number of particles. Second, it is very difficult to approximate high dimensional distribution well with a limited number of point mass approximation.

\section{Summary}
\label{sec:summary}
I proposed a new generic inference method based on optimization and ergodic deterministic transformations. This work provides us the very foundation of ergodic inference including: the fundamental ergodic inference principle; tractable estimation of ergodic loss and the its gradient; a generic construction of approximation family.

\nocite{langley00}

\bibliography{database}
\bibliographystyle{icml2018}

\end{document}